# Toward Automated Virtual Assembly for Prefabricated Construction: Construction Sequencing through Simulated BIM

Gilmarie O'NEILL[1], Matthew BALL[2], Yujing LIU[3], Mojtaba NOGHABAEI[2], and Kevin HAN[2]

[1] Dept. of Civil Engineering, University of Puerto Rico at Mayagüez, Calle Post, Mayagüez, PR 00681; email: gilmarie.oneill@upr.edu
[2] Dept. of Civil Construction and Environmental Engineering, North Carolina State Univ., 2501 Stinson Dr., Raleigh, NC 27607
[3] Dept. of Automation, Beijing Institute of Technology, 5 South Zhongguancun, Haidian District, Beijing 100081

**ABSTRACT**

To adhere to the stringent time and budget requirements of construction projects, contractors are utilizing prefabricated construction methods to expedite the construction process. Prefabricated construction methods require an adequate schedule and understanding by the contractors and constructors to be successful. The specificity of prefabricated construction often leads to inefficient scheduling and costly rework time. The designer, contractor, and constructors must have a strong understanding of the assembly process to experience the full benefits of the method. At the root of understanding the assembly process is visualizing how the process is intended to be performed. Currently, a virtual construction model is used to explain and better visualize the construction process. However, creating a virtual construction model is currently time consuming and requires experienced personnel. The proposed simulation of the virtual assembly will increase the automation of virtual construction modeling by implementing the data available in a building information modeling (BIM) model. This paper presents various factors (i.e., formalization of construction sequence based on the level of development (LOD)) that needs to be addressed for the development of automated virtual assembly. Two case studies are presented to demonstrate these factors.

**INTRODUCTION**

Prefabricated construction is a potential solution to increase productivity in the construction industry and address the labor shortage issue (Xie, Chowdhury, Issa, & Shi, 2018). In prefabricated construction, building components are built in manufacturing plants and shipped and assembled in construction sites. Prefabricated construction promotes efficiency, reduces delays, improves safety, and decreases rework (Ding, Li, Zhou, & Love, 2017). However, planning of shipping and assembly can be quite challenging for complex construction projects (Taghaddos, Hermann, & Abbasi, 2018). Researchers utilized technologies, such as Global Positioning Systems (GPS), Radio Frequency Identification (RFID), and Artificial Intelligence to optimize assembly planning in construction sites (Li, Shen, & Xue, 2014; Taghaddos et al., 2018). Tserng et al. (2011) proposed an algorithm to automate assembley planning for mechanical systems such as power plants' piping systems and demonstrated that their framework could significantly reduce rework and decrease any clash in the construction site. In addition, researchers developed virtual environments to visualize prefabricated assembly process and train technicians before actual assembly to even further reduce rework and detect any possible clashes beforehand (Kayhani, Taghaddos, Noghabaee, Ulrich, & Hermann, 2019; Noghabaei, Asadi, & Han, 2019).

However, prefabricated assembly is not fully automated yet and requires experts to implement the algorithms. Development of automated virtual assembly has been reported as a potential tool to fill this gap (Balali, Noghabaei, Heydarian, & Han, 2018; Noghabaei et al., 2019; Noghabaei, Heydarian, Balali, & Han, 2020), which includes the process of simulating construction tasks involving construction components, without performing physical prototyping. Therefore, this paper focuses on improving the understanding of the assembly process for onsite construction by creating a story and explanation of the assembly of four premanufactured wall installation. To automate this process, a path planning algorithm known as A* is used to optimize the movement of the construction elements during the assembly process based on their surrounding environment (e.g., whether the path is occluded by other objects or not). A modified version of A* was used to first, disassemble prefabricated elements from BIM and in the next step, reverse the order to avoid any clashes. The developed framework takes BIM as an input and outputs the assembly animation. In addition, using this technique, the system can show the 4D BIM using as-planned models without requiring any additional labor to create assembly animations. To demonstrate the feasibility of the proposed system, the algorithms were tested in four automated assembly scenarios. Furthermore, the algorithms were tested and verified with different A* configurations and the algorithm timing orders were compared. The results show the effectiveness and applicability of the system in automated assembly tasks.

## BACKGROUND

### Automated Assembly

Computer modeled systems can avail the visualization of different tasks and the assembly process, reduce risks, delays, and improve many more areas in construction. Scott Howe (2000) discussed the lack of knowledge in the automated construction process especially in the design industry. Meanwhile, research and development in automation for project management has been a direct influencer in the construction process. On the other hand, AbouRizk et al. (2011) explained how consequential it is to have designers, contractors and constructors understand the long-term simulation concepts in the construction area. The more these parties know about the process the easier it is to design and construct a project. Advancement in automation is going to evolve and extend the usage of different simulation programs rather than just providing general tools. A strong automation simulation framework can enable the construction industry to benefit from understanding project simulation concepts. To develop an automated simulation framework we must first understand the detail and the type of components associated with the model.

### LOD Format

Level of Development (LOD) is a scale of specifications describing how developed a model is. LOD indicates the reliability and specificity of information in construction modeling as Hooper (Hooper, 2015) explained. The LOD of a model is described using levels from LOD 100 to LOD 500. LOD 100 has the least amount of detail, and LOD 500 has the most detail. However, LOD is not only used to describe a model, but it can also be used in project management to define how reliable and detailed information is over the duration of the completed process (Zhao, Kam, T. Y. Lo, Kim, & Fischer, 2018). This research is focused on applying automated assembly simulation to LOD 300 models with prefabricated components.

**Prefabricated Materials**
The manufacturing industry has seen a significant increase in productivity relative to the construction industry. This increase has resulted from the ability to benefit from a factory manufacturing setting versus a field construction setting. The construction industry is taking advantage of the advancements in the manufacturing industry by utilizing prefabricated materials. These prefabricated materials must be installed onsite. Therefore, to help understand the assembly process on-site, this paper develops an example assembly of premanufactured walls. Taking into consideration the precast construction process Chen et al. (2016) developed a study to understand the benefits of the assembly process. They used a research programmer to create an automated premanufactured construction simulation and increased the understanding of the entire assembly process. As the process of assembling, construction, and project management becomes easier with premanufactured materials, more companies and people will see the advantages and develop it further. Understanding the assembly of premanufactured materials is important to consider the disadvantages of it. Anvari et al. (2016) developed a method for the manufacturing, transportation, and assembly of precast materials. A problem they encountered was it requires spending more time on preparation. Furthermore, not everyone is prepared to schedule the whole process of premanufactured construction and selecting the different kinds of resources needed. A simulation of this process will help companies understand the process better and be more confident in the decisions they make.

**A\* Path Planning Algorithm**
An assembly simulation at its basic necessity is a path planning exercise. Path Planning is creating an algorithm configuration process to solve problems in different environments (Beard & McLain, 2015). Reddy (2013) gives an extensive explanation of the A\* algorithm, by inserting a formula to find the best and shortest path to the goal and also to manipulate it while developing the algorithm and its restrictions. The formula is $f(x) = g(x) + h(x)$, g(x) being the distance between the start node and the current node and h(x) being the heuristic that will influence the process of the algorithm providing an approximation of the distance from the current node to the target node.

Different researchers approach it differently because of its wide range in application. Wang et al. (2015) compared the development of automated guided vehicles with the A\* path planning algorithm to find the shortest time path to complete the process. This paper focuses on determining the shortest path of least resistance to assemble the components as efficiently as possible.

**ASSEMBLY PROCESS**

Design is an ill-structured process that does not contain a direct path to its intended destination (Abou-Ibrahim & Hamzeh, 2016). The proposed approach simulates virtual assembly with BIM models to determine an efficient path. Different LOD BIM models lead to diverse construction logic/sequence. In order to develop an automated virtual assembly with different construction logics, a literary review was conducted on the background of constructing walls. To develop this idea, a step-by-step construction scenario of the assembly process was created and focused on LOD 300 models. Then, considering these steps, create an assembly and disassembly algorithm to automate the construction process. The virtual assembly is portrayed in Unity 3D using the path finding algorithm A*.

**LOD 300**

Level of Development (LOD) 300 models are good at modeling elements with a specific location, orientation, quantity, size, and shape. Abou-Ibrahim and Hamzeh (2016) discussed how with this level of development a BIM model contains enough detail that concrete beam can have specific dimensions and can be checked for deflection under loading, environmental compatibility, and different technical specifications but still lacks necessary components to put the model into service. LOD 300 models contain the necessary components without cluttering the model with too much detail. All the major components in a project are visualized, and therefore a simulation of critical components can be developed. Our goal in LOD 300 is to understand an overview of how the key components can be arranged and scheduled in the most efficient way for general construction.

*Wall assembly process*

A level and square floor platform was prepared prior to the wall construction with the substructure and foundation material in place. Then, all the walls were transported and placed around the construction area in their specified positions (see Figure 2). Wall #1 was assembled on the platform in its indicated position by using a crane. Its long axis was rotated from lying in the horizontal x-axis to the vertical z-axis and placed onto the foundation. Temporary wall braces were installed to support the wall.

After wall #1 was in place, wall #2 was rotated from the x-axis to the z-axis onto the foundation and temporary braces were installed. The 90-degree angle between wall #1 and wall #2 was verified. As shown in the chart below, the same process is repeated until wall #4 is in place (see Figure 1). Then all the temporary braces were disassembled from the walls. A final verification that all corners form a 90-degree angle was performed. The wall assembly process shown in the illustration below (see Figure 2) most closely mimics what can be expected from a simple LOD 300 model and is similar to the wall assembly process we automated.

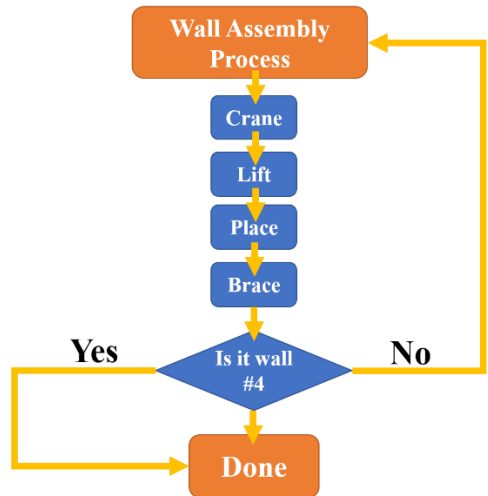
**Figure 1. LOD 300 wall assembly process**

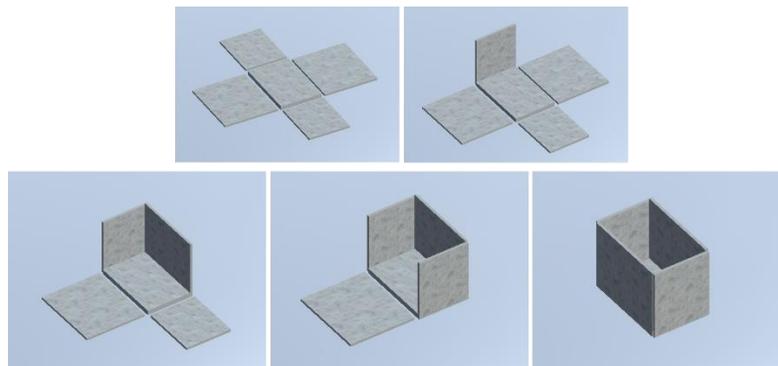
**Figure 2. Unity 3D wall assembly process**

**Implement A* path finding algorithm**

As mentioned in the wall construction assembly, when creating an algorithm, different restrictions and specifications are required depending on the size or type of components being assembled. The following path-finding algorithm was created to simulate both assembly and disassembly situations. There are four steps for implementing the A* algorithm in the assembly situation (see Figure 3).

First, a grid was setup to create all the available nodes in our world environment. Next, the heuristic (h) value of the formula was modified and marked whether each node was walkable. For this step, three aspects need to be taken into consideration.

- The Physics function in Unity detects the nodes occupied by an object on the plane. These nodes are assigned a high h value to mark the node as unwalkable
- The unoccupied nodes are then checked to see which ones will cause a collision between walls if the to-move wall is on the node. The number of nodes a wall occupies on either side of its center is stored, and the Physics function is used to determine if a collision occurs as the wall moves to a specific node. If a collision occurs at a node, the node is assigned a high h value.

- All the other nodes are safe to move on. They are assigned a low h value to mark them as walkable.

Third, start A* pathfinding. Beginning with the start node, the algorithm will select the neighbor (collection of nodes) with the lowest f value before each step. Note, since we already modified the h value of each node and we know the formula of calculating f we can easily predict that the algorithm will automatically avoid selecting the neighbor node with a high h value. Fourth, if the neighbor node being moved to is the target node, the algorithm has successfully found the path. End the A* algorithm and store the nodes generated by A* into the list path for future use.

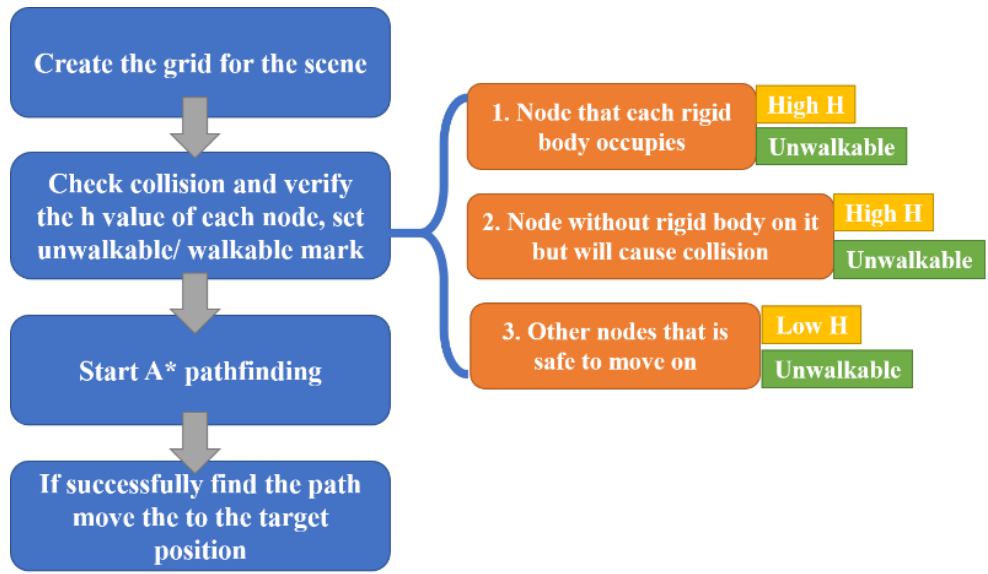

**Figure 3. Assembly process with A* path finding algorithm**

*Disassemble process*

The above algorithm can successfully generate a path for each wall regardless of whether there are any unwalkable nodes in the environment. We can apply the algorithm in the situation like assembly when we assemble the walls one by one and already know the correct order. However, this algorithm will fail in the disassembly situation because there may be some unwalkable nodes in the path calculated as a result. For example, wall A is blocked by wall B and wall C, we have to move wall B and C first before we move wall A.

The method we used to solve this problem was to create a queue at the very beginning to store all the walls we want to move in a sequence, note that the sequence may not be identical for moving without collision, but the algorithm will go through the walls stored in the queue in sequence while running. Every time we visited the first wall stored in the queue and get the path generated by A*, if there is any node marked unwalkable in the path list, which means we cannot move this wall right now, we deleted this wall member from the queue and add it to the last member of the queue.

An additional step needs to be added when setting the heuristic value. The original algorithm will mark the nodes a wall currently occupies as unwalkable. This will result in every wall being unable to move. For every process, we moved the wall to the target position and deleted this wall member from the queue decreasing the number of the walls in the queue. The algorithm will continue running until the number of walls in the queue is zero, meaning all the walls we wanted to move are in their target position. The whole process is shown in the picture below (see Figure 4). With this process, we are able to predict the right sequence for moving all the walls. For example, in the figure below (see Figure 5), the wall marked with yellow stripes was stored as the 3$^{rd}$ wall the original sequence of the queue, but it will cause a collision with the dotted wall if we don't move the white dotted wall from its original position. Following the process of the algorithm, we will skip the striped wall first by deleting it from the current position and add to the last position of the queue to see if we can generate an identical path for it later.

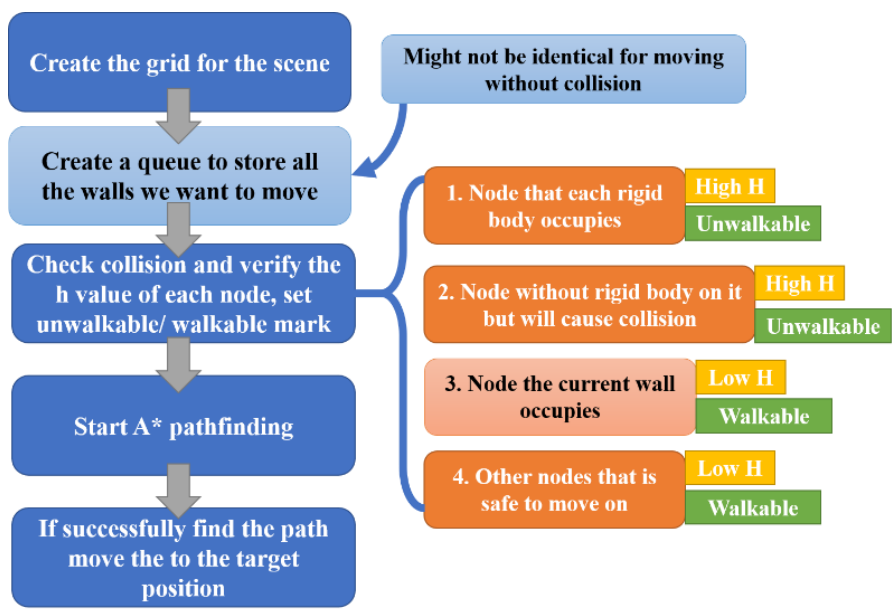

**Figure 4. Disassembly process with A* path finding algorithm**

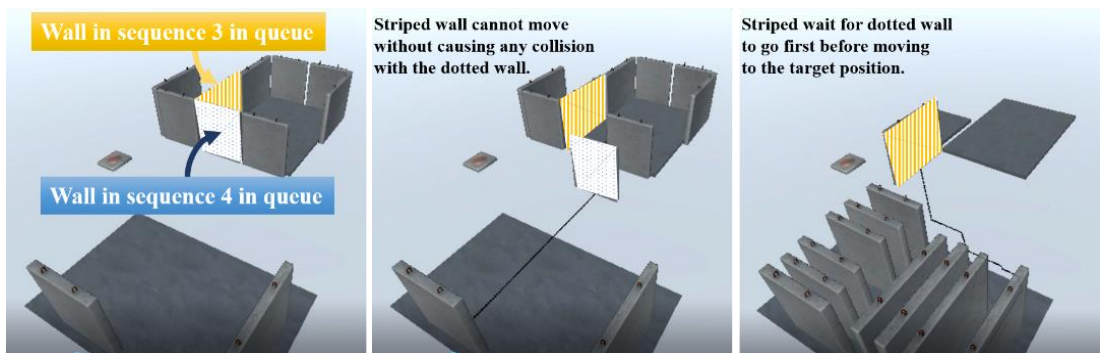

**Figure 5. Predicting the right sequence for the wall disassembly**

To test the performance of our algorithm, the total time it took to move all the walls in the different scenarios was recorded (see Table 1). There are mainly 2 factors that will influence the final performance of the algorithm. The first is the number of walls that were designed to move in different situations. It's easy to see that the time for moving all the walls will be longer with more walls stored in the queue (see Figure 6). The second factor is the radius of the node representing how big a single node can be set. A smaller radius node will increase the number of nodes on the grid. Thus, in turn, will require more calculations, comparing, and a longer selecting process during the A* algorithm when predicting the path, resulting in longer time for the whole process (see Table 1). However, a very large node will restrict the movement of the walls and may result in a solution that contains collisions between walls. Our study used various sizes of nodes to determine the least computationally expensive node size that still yielded a successful solution for the given model.

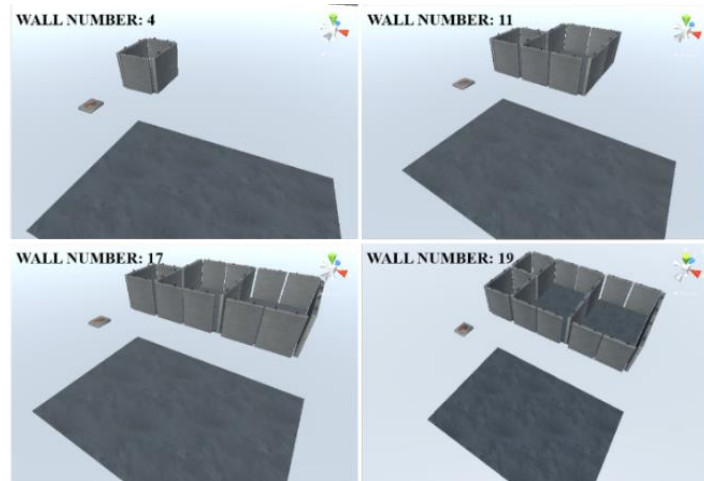

**Figure 6. Construction site for different number of walls**

**Table 1. Wall Number vs. Grid Radius**

| Grid Radius (meter) | Number of Walls | | | |
| --- | --- | --- | --- | --- |
| | 9 | 11 | 17 | 19 |
| 0.5 | 55.81 s | 166.51 s | 328.44 s | 500.52 s |
| 1 | 17.24 s | 48.34 s | 86.02 s | 117.29 s |
| 1.5 | 13.50 s | 32.22 s | 60.06 s | 75.50 s |
| 2 | 10.85 s | 24.70 s | 40.94 s | 51.43 s |
| 2.5 | 11.86 s* | 28.63 s* | 46.14 s* | 58.14 s* |
| 3 | 11.52 s* | 26.58 s* | 44.53 s* | 58.59* |

\* Collision of a moving wall and another component

**FUTURE WORK**

Currently our automated virtual assembly model is a proof of concept model that will be expanded to improve its abilities. The A* algorithm will be exchanged for a more robust algorithm and computer vision elements will be added to increase the accuracy and efficiency of the algorithm. The complexity of our testing BIM model will be increased, so we can achieve automated assembly with an equivalent LOD 400 or greater model. When this level of automation is achieved, the algorithm will become a viable tool for the construction industry.

## CONCLUSION

Technology is advancing in an expeditious manner, contractors, designers, constructors, and other participants in the construction process need to advance with it. This paper reveals an efficient methodology to assemble prefabricated walls utilizing virtual simulation with the development of BIM technology. Compared with the traditional on-site construction methods, prefabricated materials with good quality control will enable better manufacturability and assembly on-site. To further support and achieve quality assurance and plan for the assembly process, a step-by-step process utilizing a LOD 300 model was created. This step-by-step process demonstrates how typically the assembly process is understood by someone in the industry. To understand and visualize this process better, an automated virtual assembly model was created using A* path finding algorithm. A* can integrate expert knowledge on how to visualize the quality assurance path for the construction process. This method improves the efficiency of the traditional on-site assembly analysis and optimizes the process for the constructors.

## ACKNOWLEDGEMENT

This work was partially funded under Versatile Test Reactor Program of the U.S. Department of Energy (DOE) by Battelle Energy Alliance, LLC (BEA), the Management and Operating Contractor of Idaho National Laboratory (INL) under Contract Number DE-AC07-05ID14517 with DOE. The United States Government has a non-exclusive, paid-up, irrevocable, world-wide license to publish and reproduce the published form of this work and allow others to do so, for United States Government purposes. Any opinions, findings, conclusions and/or recommendations expressed in this paper are those of the authors and do not reflect the views of the entities above.